\documentclass{article}
\usepackage[preprint]{neurips_2023}

\usepackage[utf8]{inputenc} 
\usepackage[T1]{fontenc}    
\usepackage{hyperref}       
\usepackage{url}            
\usepackage{booktabs}       
\usepackage{amsfonts}       
\usepackage{nicefrac}       
\usepackage{microtype}      
\usepackage{xcolor}         
\usepackage{color}
\usepackage{graphicx}
\usepackage{amsmath}
\usepackage{amssymb}
\usepackage{verbatim}
\usepackage{caption}
\usepackage{multirow}
\usepackage{bm}
\usepackage{siunitx} 
\usepackage{comment}
\usepackage{wrapfig}
\usepackage{float}
\usepackage[super]{nth}
\usepackage{listings}
\usepackage{xparse}
\usepackage{enumitem}
\usepackage{paralist}
\usepackage{enumerate}

\newtheorem{task}{Task}

\newcommand{\etal}{et al. }

\title{Segment Anything in Non-Euclidean Domains: Challenges and Opportunities}

\author{
Yongcheng Jing$^1$,
Xinchao Wang$^2$,
Dacheng Tao$^1$
\vspace{1mm}
 \\ 
$^1$School of Computer Science, The University of Sydney, Darlington, NSW 2008, Australia,\\
$^2$Department of Electrical and Computer Engineering, National University of Singapore, Singapore\\
{\tt
xinchao@nus.edu.sg,
dacheng.tao@gmail.com
}
}

\begin{document}

\maketitle

\begin{figure}[H]
  \centering
  \includegraphics[width=\textwidth]{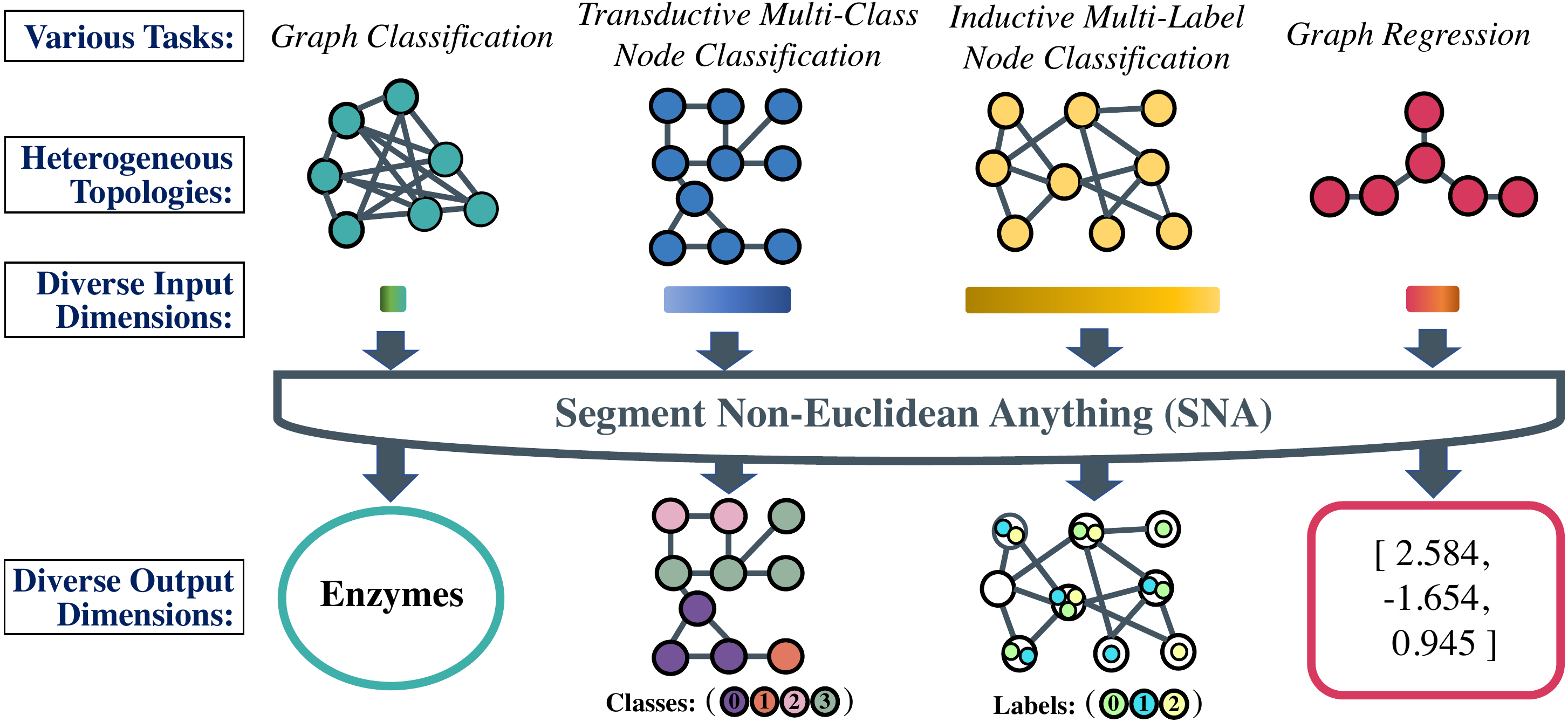}
  \caption{Illustrations of proposed foundation model in the non-Euclidean domain, which is referred to as ``\emph{Segment Non-Euclidean Anything (SNA)}''.}
  \label{fig:intro}
\end{figure}

\begin{abstract}
The recent work known as \emph{Segment Anything (SA)} has made significant strides in pushing the boundaries of semantic segmentation into the era of \emph{foundation models}.
The impact of SA has sparked extremely active discussions and ushered in an encouraging new wave of developing foundation models for the diverse tasks in the Euclidean domain, such as object detection and image inpainting.
Despite the promising advances led by SA, the concept has yet to be extended to the non-Euclidean graph domain.
In this paper, we explore a novel \textbf{\emph{Segment Non-Euclidean Anything (SNA)}} paradigm that
strives to develop foundation models that can handle the diverse range of graph data within the non-Euclidean domain, seeking to expand the scope of SA and lay the groundwork for future research in this direction.
To achieve this goal, we begin by discussing the recent achievements in foundation models associated with SA.
We then shed light on the unique challenges that arise when  applying the SA concept to graph analysis, which involves understanding the differences between the Euclidean and non-Euclidean domains from both the data and task perspectives.
Motivated by these observations, we present several preliminary solutions to tackle the challenges of SNA and detail their corresponding limitations, along with several potential directions to pave the way for future SNA research.
Experiments on five \emph{Open Graph Benchmark (OGB)} datasets across various tasks, including graph property classification and regression, as well as multi-label prediction, demonstrate that the performance of the na\"ive SNA solutions has considerable room for improvement, pointing towards a promising avenue for future exploration of \emph{Graph General Intelligence}.
\end{abstract}

\section{Introduction}
\label{sec:intro}

Foundation models, as named in \cite{bommasani2021opportunities}, refer to models that demonstrate strong generalization across a wide range of downstream tasks, achieved through training on large-scale data.
The era of foundation models marks a novel paradigm of artificial intelligence, which has been predominantly studied in the field of natural language processing (NLP) over the past few years. This line of research has led to several ground-breaking works, such as BERT \cite{devlin2018bert}, GPT-3 \cite{brown2020language}, and GPT-4 \cite{openai2023gpt4}. 
Of particular interest is the recent emergence of ChatGPT \cite{chatgpt}, a specific application of the \emph{Large Language Model (LLM)} of GPT-3.5. ChatGPT's impressive ability to align natural language with human intent through nteractive conversations has sparked a revolution towards artificial general intelligence \cite{zhou2023comprehensive}.

Recently, the field of computer vision also witnessed a surge in the popularity of learning task-agnostic foundation models \cite{wang2023seggpt,oquab2023dinov2,yu2023inpaint,ShilongLiu2023GroundingDM,chen2023semantic,xu2023semantic,zou2023segment,wang2023seggpt,fang2023depgraph,yang2022deep}.
This movement, which we term as ``Artificial Intelligence for Anything'', was sparked by the seminal work of Kirillov \etal \cite{kirillov2023segment} with their \emph{Segment Anything} (SA) model.
SA introduced a prompt-based approach to segmentation that demonstrated impressive generalization capabilities across a range of downstream segmentation tasks.
Concurrently, Zou \etal \cite{zou2023segment} explored the use of a promptable, interactive model named SEEM (\emph{Segmenting Everything Everywhere}) that offers support for a wider range of high-level semantic tasks and prompt types than SA. Notably, SEEM enables the composition of all prompt types, the complete coverage of every pixel in an image, and the semantic segmentation of every object.
Additionally, Wang \etal \cite{wang2023seggpt} also contributed to this field with the development of SegGPT, a generalist model inspired by the visual prompt paradigm \cite{bahng2022exploring,bar2022visual} that achieves zero-shot segmentation tasks. 
These advancements demonstrate significant promise for the development of robust and versatile foundation models for computer vision applications.

The breakthrough in visual foundation models has rapidly catalyzed further exploration in the field, leading to the development of several projects aimed at extending the SA paradigm.
These include
\emph{Paint-Anything} \cite{liu2023paint}, \emph{Transfer-Any-Style}\cite{liu2023transfer,liu2013anythingstyle}, \emph{Anything-3D} \cite{yang20233d,shen2013anything3d}, \emph{Inpaint-Anything} \cite{yu2023inpaint}, \emph{Napari-Segment-Anything} \cite{napari2023seg}, \emph{Image2Paragraph} \cite{show2023image}, \emph{Segment-and-Track-Anything} \cite{yang2023seg} and \emph{Caption-Anything} \cite{wang2023caption}.
A comprehensive survey of these developments is provided by Fang \etal in \cite{fang2023awesome}.

Despite the inspiring progress of visual foundation models in the Euclidean domain, there remains a lack of research studying foundation models in the non-Euclidean graph domain. To address this gap, this paper takes, to our knowledge, the first step towards developing foundation models tailored for non-grid graph data in the non-Euclidean domain, which we term as the ``\emph{Segment Non-Euclidean Anything (SNA)}'' paradigm. The goal is to construct a foundation model for graph-related tasks, such as node- and graph-level property prediction and edge prediction, capable of handling diverse and irregular graph data with varying node numbers and feature dimensions across different domains, as shown in Fig.~\ref{fig:intro}.

In terms of data, images are structured as regular grids with fixed three channels as inputs, while graphs exhibit irregular structures with diverse input channels and various topological configurations. Graphs also exhibit a higher degree of diversity than images, ranging from homogeneous graphs with single node and edge types to heterogeneous graphs containing various nodes and links across different domains such as social networks, molecule graphs, traffic graphs, and e-commerce product knowledge graphs. On the task side, graph analysis encompasses various task levels and settings, including transductive and inductive scenarios, and graph, node, and edge-level representation learning. The diversity of graph structures, data types, and tasks poses significant challenges to the development of foundation models in the non-Euclidean domain.

Driven by these observations, we propose several novel methods to alleviate the challenges faced by SNA. Specifically, we propose an meta-slimmable graph convolutional operation to address the issue of diverse input dimensions. This operation adaptively identifies a corresponding subset of neurons to handle the varying-dimension graphs. Furthermore, to tackle the problem of diverse-domain graph data and distinct task levels and settings, we propose the idea of a prompt-based graph analysis framework. This framework enables our SNA model to adapt to various downstream graphs and tasks by accepting an example labeled input-output pair as inputs. We acknowledge that our proposed methods are currently in a preliminary stage and intend to refine them further with feedback from the research community.

In sum, this paper makes four contributions to the field of foundation models:
\begin{compactitem}
\item We give an overview on the evolution of foundation models in the Euclidean domain, with a focus on the recent advancements in foundation models associated with \emph{Segment Anything};
\item We provide a discussion on the existing works related to foundation models in the non-Euclidean graph domain;
\item We propose the task of \emph{Segment Non-Euclidean Anything}, identify major challenges, and present several na\"ive SNA methods to address the dilemma;
\item We present preliminary experimental results that demonstrate the inadequacy of existing vanilla solutions, highlighting the need for improved methods in the non-Euclidean domain.
\end{compactitem}
We hope that our work can inspire future research into developing foundation models in the non-Euclidean domain.

\section{Related Work: \emph{The Evolution of Foundation Models}}
\label{sect:foundation}

In this section, we provide a brief overview of the development of foundation models in the Euclidean domain, spanning from their origins, their significant rise following the introduction of SA, and the new era of ``\emph{visual intelligence for anything}''.

\subsection{The Origins of Foundation Models}

The advent of\emph{Natural Language Processing (NLP)} in late 2018 inaugurated the era of foundational models,
epitomized by the proliferation of \emph{Large Language Models (LLMs)} including but not limited to DALL-E \cite{ramesh2021zero}, BERT \cite{devlin2018bert}, and GPT-3 \cite{brown2020language}. These models are capable of adapting to diverse downstream tasks, thereby demonstrating their remarkable versatility. The first generation of foundation models has demonstrated an impressive range of linguistic capabilities, exhibiting a surprising degree of adaptability across a wide spectrum of linguistic contexts.

The burgeoning success of foundational models in NLP has sparked a wealth of research in this field. One such comprehensive investigation is presented by Bommasani \etal \cite{bommasani2021opportunities}, which delves into the technical aspects of foundation models, including their model architectures, training procedures, data, systems, security, and theory, as well as their societal impact, specifically concerning their potential for inequity and misuse.
The study draws attention to the opportunities and risks associated with foundation models, which are based on conventional deep learning and transfer learning methodologies but possess emergent capabilities due to their scale. Furthermore, Bommasani \etal highlight the challenges of utilizing foundation models for lengthy legal documents and narratives. \cite{bommasani2021opportunities} also suggests ways to assess the efficacy of foundation models and minimize the potential risks associated with their use.
Bommasani \etal acknowledge that foundational models are still shrouded in uncertainty and that the current research marks only the beginning of a paradigm shift in the realm of AI systems.

The concept of foundational models in computer vision was introduced very early in the work of Hu \etal \cite{hu2018learning} titled ``\emph{Learning to Segment Every Thing}''. This pioneering study, published in the early stages of research on visual foundation models, paved the way for the development of more advanced and accurate models in computer vision.
Specifically, Hu \etal propose novel methods for the partial supervision training and weight transfer function of Mask R-CNN models to attain object instance segmentation. The work of  Hu \etal addresses the challenge of using a smaller amount of annotated data for large-scale training and aim to surpass the limitations of current instance segmentation models, which are typically limited to 100 object categories. The new transfer learning approach leverages mask prediction and task transfer learning strategies to construct a model that can efficiently segment classes with only bounding box annotations for local portions. Through this approach, each detected object can be segmented into foreground and background parts, leading to accurate object instance segmentation.

\begin{table}[t]
  \renewcommand\arraystretch{1.1}
  \setlength\tabcolsep{10 pt}
  \caption{An overview of the extensions of \emph{Segment Anything}.}
  \begin{center}
  \begin{footnotesize}
  \begin{tabular}{c|c}
      \noalign{\hrule height 0.8pt}
      \textbf{Paper / Repository} &  \textbf{Title}\\
      \hline
      \cite{fang2023awesome} & Awesome-Anything \\ \hline
      \cite{jiang23awesome} & Awesome Segment Anything \\ \hline
      \cite{xu23awesome} & Awesome-Segment-Anything-Extensions \\ \hline
        \cite{chen2023semantic} & Semantic Segment Anything \\ \hline
        \cite{xu2023semantic} & Segment and Track Anything \\ \hline
        \cite{yang2023seg} & Segment-and-Track-Anything \\ \hline
        \cite{wang2023caption} & Caption Anything \\ \hline
        \cite{liu2023paint} & Paint Anything \\ \hline
        \cite{liu2023transfer} & Transfer Any Style \\ \hline
        \cite{yang20233d} & Anything-3D \\ \hline
        \cite{show2023image} & Image.txt: Transform Image Into Unique Paragraph \\ \hline
        \cite{napari2023seg} & Napari-Segment-Anything  \\ \hline
        \cite{yu2023inpaint} & Inpaint Anything  \\ \hline
        \cite{napari2023seg} & Napari-Segment-Anything \\ \hline
        \cite{idea23grounded} & Grounded-Segment-Anything \\ \hline
        \cite{cao23grounded} & GroundedSAM-Zero-Shot-Anomaly-Detection \\ \hline
        \cite{fei23image} & Image Editing Anything \\ \hline
        \cite{sea23edit} & Edit Anything by Segment-Anything \\ \hline
        \cite{rocky23prompt} & Prompt-Segment-Anything \\ \hline
        \cite{li23sam} & SAM-RBox  \\ \hline
        \cite{tin23open} & Open-Vocabulary-Segment-Anything  \\ \hline
        \cite{zhao23seg} & SegDrawer \\ \hline
        \cite{han23sam} & Segment Anything EO Tools \\ \hline
        \cite{viet23sam} & AnyLabeling  \\ \hline
        \cite{yang23sam} & Optical Character Recognition with Segment Anything (OCR-SAM)  \\ \hline
        \cite{amine23sam} & SAM Medical Imaging \\ \hline
        \cite{jay23sam} & LIME-SAM \\ \hline  
        \cite{liu23sammm} & SAM + MMDetection  \\ \hline
        \cite{dv23sam} & 3D-Box via Segment Anything \\ \hline
        \cite{gao23sam} & Track-Anything \\ \hline  
        \cite{achal23sam} & Towards Segmenting Anything That Moves (Segment-Any-Moving)  \\ 
        \noalign{\hrule height 0.8pt}

       \end{tabular}
       \end{footnotesize}
  \end{center}
  \footnotesize
  \smallskip
  \label{table:anythinganything}
  \end{table}

\subsection{The Catalyst for the Rise of Foundation Models: \emph{``Segment Anything''}}

Despite the novel concept of the visual foundation model proposed in \cite{hu2018learning}, its segmentation performance has shown limitations, failing to draw significant attention to the research of vision foundation models. However, in a recent preprint titled `\emph{Segment Anything (SA)}'' \cite{kirillov2023segment}, Kirillov \etal present an improved model that achieves more encouraging and impressive performance. This work has garnered significant attention from the computer vision research community towards the field of visual foundation models.

The SA project in \cite{kirillov2023segment} aims to establish a novel foundation model for image segmentation by leveraging a promptable segmentation model, and an extensive collection of over 1 billion masks. The proposed model design and training approach can be transferred to various tasks with zero-shot learning, demonstrating impressive zero-shot performance on multiple tasks, including competitive or superior performance to supervised models.
The paper of \cite{kirillov2023segment} also discusses the challenges of image segmentation, including tasks, models, and data, and presents corresponding solutions. Notably, the authors have created the SA-1B dataset, consisting of over 1 billion masks and 110 million images, which is openly available for future research. The SA project has significant implications for advancing the field of image segmentation, particularly in the realm of zero-shot task-agnostic learning of vision foundation models.

A concurrent work by Zou \etal \cite{zou2023segment} also  developed a foundational segmentation model, referred to as SEEM, which aims to support various segmentation tasks and multiple interactive modes. The proposed model in \cite{zou2023segment} utilizes learnable memory prompts and hierarchical segmentation to enhance semantic perception, demonstrating strong generalization ability and efficient computational performance. Experimental results demonstrate SEEM's effectiveness across multiple segmentation tasks, with a larger range of prompt types compared to the previously proposed SA model.

Meanwhile, the work in \cite{wang2023seggpt} also presents a universal segmentation model, termed as SegGPT, that simultaneously addresses several segmentation tasks, including arbitrary object segmentation, multi-part segmentation, iris segmentation, video object segmentation, and finite-semantic segmentation, within a single framework. The devised approach in \cite{wang2023seggpt} uses random color mapping and contextual examples for training instead of relying on fixed colors, making the model more flexible and adaptive. 
As a result, the proposed SegGPT in  \cite{wang2023seggpt} exhibits strong performance in both in-domain and out-of-domain segmentation tasks, and can generalize effectively across different domains, thereby enabling automatic execution of various segmentation tasks and providing a universal solution for segmentation problems.

\subsection{The New Era of Foundation Models: \emph{``Visual Intelligence for Anything''}}

The success of the SA project has drawn significant attention in the computer vision field and has sparked a wave of research and development in vision foundation models in Euclidean domains \cite{liang2022open,wang2023v3det,dave2019towards,xu2022pose,shen2023hugginggpt,yang2022deep,wang2023seggpt,oquab2023dinov2,yu2023inpaint,ShilongLiu2023GroundingDM,fang2023depgraph,chen2023semantic,xu2023semantic,zou2023segment,wang2023seggpt,yang2022factorizing,liang2023taskmatrix}. As shown in Tab.~\ref{table:anythinganything}, recent extensions of the SA approach span a broad range of domains, such as \emph{Image.txt: Transform Image Into Unique Paragraph} \cite{show2023image}, \emph{AnyLabeling} \cite{viet23sam}, \emph{Anything-3D} \cite{yang20233d}, \emph{Caption-Anything} \cite{wang2023caption}, \emph{Paint-Anything} \cite{liu2023paint,liu2021paint},  \emph{Inpaint-Anything} \cite{yu2023inpaint}, \emph{Transfer-Any-Style}\cite{liu2023transfer,liu2021adaattn}, \emph{Napari-Segment-Anything} \cite{napari2023seg}, and \emph{Segment-and-Track-Anything} \cite{yang2023seg}.

An especially noteworthy development is the recent work known as DINOv2 \cite{oquab2023dinov2}, which presents a novel unsupervised learning approach for producing versatile visual features that can be utilized across a wide range of image distributions and tasks. The method proposed in \cite{oquab2023dinov2} employs self-supervised learning and extensive data preprocessing to construct filtered and balanced datasets. The DINOv2 model, which is trained using the feed-forward neural network Vision Transformer architecture, demonstrates exceptional quality in various standard computer vision benchmark tests, thus highlighting the effectiveness of this simplified technique for visual reasoning.

Despite the growing interest in the development of foundation models in the field of computer vision, there remains a significant gap in the literature regarding the exploration of such models in the non-Euclidean graph domain. In response, this study aims to address this gap by extending the concept of foundation models to the domain of graph analysis. The primary goal of this work is to investigate the feasibility and effectiveness of incorporating foundation models in the context of non-Euclidean graphs, thereby contributing to the advancement of foundation models in this area.

\section{Segment Non-Euclidean Anything Task}

\subsection{Preliminaries}

Graph neural networks (GNNs) have emerged as a powerful model for non-grid data, where data is represented as a set of nodes with associated features, and the relationships among the nodes are represented as a graph \cite{kipf2017semi,wu2020comprehensive,zhou2018graph,dwivedi2020benchmarking,jing2022learning,yan2018spatial,li2018adaptive,jing2021meta,yang2020factorizable,ma2019disentangled,wang2018graphgan,yang2021spagan,huang2018adaptive,li2018deeper,yang2020distillating,nt2019revisiting}. The first-order approximation of spectral graph convolutions forms the basis of Graph Convolutional Network (GCN) \cite{kipf2017semi}. A range of methods have been proposed to enhance the performance of GCN, such as GraphSAGE by Hamilton \etal \cite{hamilton2017inductive}, which makes the model scalable by sampling neighbors instead of using all of them. Graph attention network (GAT) \cite{velivckovic2018graph} introduces the attention mechanism to learn the weights for each neighbor automatically. Other techniques like adaptive sampling have been proposed for efficient training \cite{huang2018adaptive}. In recent years, several novel techniques such as PairNorm layer \cite{zhao2020pairnorm} and DropEdge \cite{rong2020dropedge} strategy have been proposed to overcome the oversmoothing problem in GNNs \cite{rusch2023survey}.

Despite the encouraging progress in graph neural networks, to the best of our knowledge, there is little research towards learning task-agnostic foundation models for unversal graph analysis.
But indeed, there are some works that are related to this topic of graph foundation models, such as knowledge transfer and graph reprogramming.
In particular, the work in \cite{jing2021amalgamating} introduces a novel knowledge transfer tasks that aims to obtain a multi-talented and lightweight student model, by amalgamating knowledge from heterogeneous teacher GNNs.
This is specifically achieved by the use of topology attribute mapping (TAM) to clarify the topological semantics across different networks and a slimmable graph convolutional operation to tackle diverse feature dimensions.
Also, the research presented in \cite{jing2023deep} introduces a set of deep graph reprogramming techniques. These techniques allow for the adaptation of a pre-trained GNN model to a range of downstream tasks that were not previously seen during training. This is achieved by merely modifying a portion of the model inputs. As such, this technique may be regarded as a basic form of graph foundation model.

\subsection{Task Definition}

Inspired by the advance of ``\emph{Segment Anything}'' in the Euclidean domain, we proposes in this paper a novel task, termed as ``\emph{Segment Non-Euclidean Anything (SNA)}'', that aims to develop a foundational model that can readily support diverse graph-related downstream tasks in the Non-Euclidean domain:
\begin{task}[\textbf{Segment Non-Euclidean Anything}]
  \label{task}
  \textit{The objective of Segment Non-Euclidean Anything (SNA) is to establish a foundation model within the Non-Euclidean domain for universal graph analysis. This includes handling various topological graph samples and diverse tasks, such as transductive and inductive node property prediction, link prediction, and graph classification and regression.}
\end{task}
The target \emph{Segment Non-Euclidean Anything Model (SNAM)} is expected to have the following merits:
\begin{compactitem}
\item[$\blacktriangleright$] 
SNAM has the ability to handle diverse graph samples from various domains, including citation networks, e-commerce product graphs, social networks, and molecule graphs, while accommodating different node/edge numbers and heterogeneous input dimensions;
\item[$\blacktriangleright$] SNAM's versatility allows it to tackle various homogeneous and heterogeneous graph analysis tasks and settings, such as transductive and inductive node classification, link property prediction, and graph classification and regression;
\item[$\blacktriangleright$] SNAM's flexibility extends to accepting various types of prompts in an interactive manner, making it easy to complete the relevant tasks.
\end{compactitem} 

\subsection{Challenge Identification}

Task~\ref{task} poses two primary challenges that must be addressed in order to achieve the ambitious goal of NSM:
\begin{compactitem}
  \item[$\divideontimes$] 
\textbf{Data Side:} 
The first challenge faced in Task~\ref{task} pertains to the management of downstream features of varying dimensions, as typical GNNs can only handle a single input dimension. For instance, nodes in the \emph{Citeseer} citation network possess 3703 input features, while those in \emph{Amazon Co-purchase} graphs have only 767. Furthermore, the output dimensions of different graph samples are also heterogeneous. In addition, the simultaneous processing of homogeneous and heterogeneous graph samples adds to the complexity of the task;
\item[$\divideontimes$] 
\textbf{Task Side:} 
The diversity of graph-related tasks poses a significant challenge for the development of a foundational SNAM for general graph analysis.
Basic graph-related tasks, which include graph-, node-, and edge-level tasks, pose specific challenges for analysis. These tasks encompass a range of objectives, including graph property classification and regression, transductive and inductive node property prediction, as well as edge classification. The varying nature of these tasks necessitates a flexible and adaptable SNAM capable of accommodating different types of graph-related objectives.
\end{compactitem} 

\subsection{Methodology Proposal}

\paragraph{Meta-Slimmable Graph Processing.}
To address the challenge of accommodating diverse feature dimensions for varying tasks, the approach proposed in \cite{jing2021amalgamating} introduces a specialized slimmable graph convolutional layer. This layer enables adaptive activation or deactivation of its channels based on the input feature dimensions. Specifically, prior to training, the slimmable graph convolutions in \cite{jing2021amalgamating} require a maximum channel number to establish the weights' shape in GCN layers. Then, given input nodes with different feature dimensions, the model dynamically utilizes the part of neurons to process the input features. By substituting the first layer and output layer with slimmable graph convolutional layers, the resulting GNN model can simultaneously process graph samples with varying input feature dimensions and different output dimensions.

While \cite{jing2021amalgamating} has shown promising results, the manual selection of neurons for different input dimensions limits its potential as a foundation model for universal graph analysis. In light of this, we introduce an improved version of the slimmable graph convolution, named \emph{Meta-Slimmable Graph Processing}. This approach employs a meta-learning strategy that learns to select the optimal neurons based on the downstream tasks, rather than relying on manual selection. By doing so, the model capacities can be increased, enabling the handling of a broader range of unseen tasks. This proposed method represents a step towards developing more versatile and adaptable foundation models for universal graph analysis.

\paragraph{Prompt-based Foundation Framework.}
Inspired by the success of the segment anything model presented in \cite{kirillov2023segment}, we introduce a prompt-based framework for general graph analysis, referred to as prompt-based SNA. This interactive framework enables users to input natural language prompts to perform a range of graph-related tasks. However, unlike the segment anything model, which mainly focuses on segmentation tasks, graph analysis tasks are more intricate and diverse, encompassing predictions at edge, node, and graph levels. To address these heterogeneous tasks, the SNA framework takes an exemplar small graph along with corresponding input and output pairs as inputs and leverages this information to replicate the necessary behavior to perform the desired downstream tasks on various input graphs. We aim to further investigate and refine the details of the proposed framework in our future work.

\section{Experiments}

In this section, we perform preliminary experiments to 
to showcase the restrictions of current vanilla methods in relation to SNA.

\paragraph{Implementation Details.} 
We employ five datasets from the Open Graph Benchmark (OGB) \cite{hu2020open} for molecular property prediction. These datasets include \texttt{ogbg-molbace}, \texttt{ogbg-molbbbp}, \texttt{ogbg-molesol}, \texttt{ogbg-molpcba}, and \texttt{ogbg-molhiv}. The \texttt{ogbg-molesol} dataset aims to perform molecular property regression, while the \texttt{ogbg-molbace}, \texttt{ogbg-molbbbp}, \texttt{ogbg-molhiv}, and \texttt{ogbg-molpcba} datasets are intended for single-label and multi-label prediction settings. Each graph within these datasets represents a molecule, with nodes denoting atoms and edges representing chemical bonds. The node features include the atomic number, chirality, and additional atom features such as formal charge.
The learning rate setting is set to 0.005, with a weight decay of $5\times 10^{-4}$.
We use the Adam optimizer for pre-training.
In this study, we utilize the slimmable graph convolutional operation, proposed in \cite{jing2021amalgamating}, as a vanilla method to adapt a pre-trained model to various downstream feature dimensions.

\paragraph{Results.}
Tab.~\ref{tab:ogb} shows the results of using the \cite{jing2021amalgamating} vanilla method to adapt a pre-trained model for downstream tasks. Despite the fact that all five OGB benchmarks used in this study are in the domain of molecule analysis, and share similar molecule graph structures as inputs, it remains highly difficult for a pre-trained model to address a wide range of tasks that were not encountered during training. These findings highlight the potential research obstacles involved in the development of foundational models for general graph analysis.
Our future work entails conducting additional experiments on a broader range of datasets and delving deeper into the underlying reasons behind our experimental results.

  \newcommand{\liuhaofive}{\fontsize{7.5pt}{\baselineskip}\selectfont}
  \newcommand{\liuhaoseven}{\fontsize{7pt}{\baselineskip}\selectfont}
  \newcommand{\liuhaoeight}{\fontsize{7pt}{\baselineskip}\selectfont}
  \begin{table}[t]
    \caption{Results on the OGB datasets across various tasks, including graph classification, property regression, and multi-label prediction.
    }
    \begin{center}
    \liuhaofive
    \setlength\tabcolsep{1.5 pt}
    {\renewcommand{\arraystretch}{0.8}
    \begin{tabular}{l|c|c|c|c|c}
      \noalign{\hrule height 0.6pt}
    
      \textbf{Pre-trained Datasets} & \multicolumn{1}{c|}{\texttt{ogbg-molbace}} & \multicolumn{1}{c|}{\texttt{ogbg-molbbbp}}& \multicolumn{1}{c|}{\texttt{ogbg-molesol}}& \multicolumn{1}{c|}{\texttt{ogbg-molpcba}}& \multicolumn{1}{c}{\texttt{ogbg-molpcba}}\\ 
      {Task Types} & \multicolumn{1}{c|}{\liuhaoseven\bf\emph{Graph Classification}} & \multicolumn{1}{c|}{\liuhaoseven\bf\emph{Graph Classification}}& \multicolumn{1}{c|}{\liuhaoseven\bf\emph{Graph Regression}}& \multicolumn{1}{c|}{\liuhaoseven\bf\emph{Multi-label Prediction}}& \multicolumn{1}{c}{\liuhaoseven\bf\emph{Multi-label Prediction}}\\ \noalign{\hrule height 0.1pt}
      {Pre-trained Results} & \multicolumn{1}{c|}{\emph{\liuhaoeight{ROC-AUC}}: 0.7734} & \multicolumn{1}{c|}{\emph{\liuhaoeight{ROC-AUC}}: 0.6709}& \multicolumn{1}{c|}{\emph{\liuhaoeight{RMSE}}: 1.3000}& \multicolumn{1}{c|}{\emph{\liuhaoeight{AP}}: 0.2145}& \multicolumn{1}{c}{\emph{\liuhaoeight{AP}}: 0.2145}\\
      \noalign{\hrule height 0.6pt}
      \textbf{Downstream Datasets} & \multicolumn{1}{c|}{\texttt{ogbg-molhiv}} & \multicolumn{1}{c|}{\texttt{ogbg-molhiv}} & \multicolumn{1}{c|}{\texttt{ogbg-molhiv}} & \multicolumn{1}{c|}{\texttt{ogbg-molhiv}} & \multicolumn{1}{c}{\texttt{ogbg-molbace}}\\
      {{Task Types}} & \multicolumn{1}{c|}{\liuhaoseven\bf\emph{Graph Classification}} & \multicolumn{1}{c|}{\liuhaoseven\bf\emph{Graph Classification}}& \multicolumn{1}{c|}{\liuhaoseven\bf\emph{Graph Classification}} & \multicolumn{1}{c|}{\liuhaoseven\bf\emph{Graph Classification}}& \multicolumn{1}{c}{\liuhaoseven\bf\emph{Graph Classification}}\\ \noalign{\hrule height 0.1pt}
      {Training from Scratch} & \multicolumn{1}{c|}{\emph{\liuhaoeight{ROC-AUC}}: 0.7530} & \multicolumn{1}{c|}{\emph{\liuhaoeight{ROC-AUC}}: 0.7530}& \multicolumn{1}{c|}{\emph{\liuhaoeight{ROC-AUC}}: 0.7530}& \multicolumn{1}{c|}{\emph{\liuhaoeight{ROC-AUC}}: 0.7530}& \multicolumn{1}{c}{\emph{\liuhaoeight{ROC-AUC}}: 0.7734}\\
      \noalign{\hrule height 0.1pt}
      {Vanilla Method} & \emph{\liuhaoeight{ROC-AUC}}: 0.4481 
       & \emph{\liuhaoeight{ROC-AUC}}: 0.3409 & \emph{\liuhaoeight{ROC-AUC}}: 0.3413  & \emph{\liuhaoeight{ROC-AUC}}: 0.5020 & \emph{\liuhaoeight{ROC-AUC}}: 0.5000 \\ 
       \noalign{\hrule height 0.6pt}
      \end{tabular}}
      {\renewcommand{\arraystretch}{0.8}
      \begin{tabular}{l|c|c|c|c|c}
        \noalign{\hrule height 0.6pt}
      \textbf{Pre-trained Datasets} & \multicolumn{1}{c|}{\texttt{ogbg-molhiv}} & \multicolumn{1}{c|}{\texttt{ogbg-molhiv}}& \multicolumn{1}{c|}{\texttt{ogbg-molhiv}}& \multicolumn{1}{c|}{\texttt{ogbg-molpcba}}& \multicolumn{1}{c}{\texttt{ogbg-molpcba}}\\ 
      {Task Types} & \multicolumn{1}{c|}{\liuhaoseven\bf\emph{Graph Classification}} & \multicolumn{1}{c|}{\liuhaoseven\bf\emph{Graph Classification}}& \multicolumn{1}{c|}{\liuhaoseven\bf\emph{Graph Classification}}& \multicolumn{1}{c|}{\liuhaoseven\bf\emph{Multi-label Prediction}}& \multicolumn{1}{c}{\liuhaoseven\bf\emph{Multi-label Prediction}}\\ \noalign{\hrule height 0.1pt}
      {Pre-trained Results} & \multicolumn{1}{c|}{\emph{\liuhaoeight{ROC-AUC}}: 0.7530} & \multicolumn{1}{c|}{\emph{\liuhaoeight{ROC-AUC}}: 0.7530}& \multicolumn{1}{c|}{\emph{\liuhaoeight{ROC-AUC}}:  0.7530}& \multicolumn{1}{c|}{\emph{\liuhaoeight{AP}}: 0.2145}& \multicolumn{1}{c}{\emph{\liuhaoeight{AP}}: 0.2145}\\
      \noalign{\hrule height 0.6pt}
      \textbf{Downstream Datasets} & \multicolumn{1}{c|}{\texttt{ogbg-molbace}} & \multicolumn{1}{c|}{\texttt{ogbg-molbbbp}}& \multicolumn{1}{c|}{\texttt{ogbg-molesol}}& \multicolumn{1}{c|}{\texttt{ogbg-molbbbp}}& \multicolumn{1}{c}{\texttt{ogbg-molesol}}\\ 
      {{Task Types}} & \multicolumn{1}{c|}{\liuhaoseven\bf\emph{Graph Classification}} & \multicolumn{1}{c|}{\liuhaoseven\bf\emph{Graph Classification}}& \multicolumn{1}{c|}{\liuhaoseven\bf\emph{Graph Regression}}& \multicolumn{1}{c|}{\liuhaoseven\bf\emph{Graph Classification}}& \multicolumn{1}{c}{\liuhaoseven\bf\emph{Graph Regression}} \\ \noalign{\hrule height 0.1pt}
      {Training from Scratch} & \multicolumn{1}{c|}{\emph{\liuhaoeight{ROC-AUC}}: 0.7734} & \multicolumn{1}{c|}{\emph{\liuhaoeight{ROC-AUC}}: 0.6709}& \multicolumn{1}{c|}{\emph{\liuhaoeight{RMSE}}: 1.3000}& \multicolumn{1}{c|}{\emph{\liuhaoeight{ROC-AUC}}: 0.6709}& \multicolumn{1}{c}{\emph{\liuhaoeight{RMSE}}: 1.3000}\\
      \noalign{\hrule height 0.1pt}
      {Vanilla Method} & \emph{\liuhaoeight{ROC-AUC}}: 0.6120  &
      \emph{\liuhaoeight{ROC-AUC}}: 0.3936 & \emph{\liuhaoeight{RMSE}}: 2.3214 & \emph{\liuhaoeight{ROC-AUC}}: 0.5268 & \emph{\liuhaoeight{RMSE}}: 3.7331 \\ 
      \hline
  
    \end{tabular}}
    \end{center}
    \label{tab:ogb}
    \end{table}

\section{Conclusions}
In this paper, we introduce a novel \emph{Segment Non-Euclidean Anything (SNA)} task, tailored for irregular and heterogeneous graph analysis.
The proposed SNA paradigm basically extends the recent \emph{Segment Anything} project to the domain of topological graph processing with the objective of building a foundation model for universal graph-related tasks.
Towards this end, we identity two primary challenges from the data and task sides, respectively, and propose two dedicated vanilla solutions, including a meta-slimmable neural operation and a prompt-inspired framework, to overcome these challenges.
Experiments on five benchmarks across various domains, including multi-label property prediction, graph classification, and property regression, demonstrate that the existing na\"ive solutions for developing a foundation graph model still have a long way to go to achieve favorable performance. 
In our future work, we will strive to refine the proposed SNA approaches through collaboration with passionate researchers in this field and pushing the boundaries towards building a well-established foundational model for graph general intelligence.

\section*{Acknowledgements}
This research is supported by Australian Research Council Projects in part by FL170100117 and IH180100002, and the National Research Foundation Singapore under its AI Singapore Programme (Award Number: AISG2-RP-2021-023).

\medskip
{
\bibliographystyle{unsrtnat}
\bibliography{MYRE}
}

\end{document}